# LEARNING POINT EMBEDDING FOR 3D DATA PROCESSING

*Chen Zhenpeng*  *Li Yuan*

Intelligent Driving Dept, GAC R&D Center, GuangZhou, China
chenzpbj@hotmail.com  liyuan@gacrnd.com

## ABSTRACT

Among 2D convolutional networks on point clouds, point-based approaches consume point clouds of fixed size directly. By analysis of PointNet, a pioneer in introducing deep learning into point sets, we reveal that current point-based methods are essentially spatial relationship processing networks. In this paper, we take a different approach. Our architecture, named PE-Net, learns the representation of point clouds in high-dimensional space, and encodes the unordered input points to feature vectors, which standard 2D CNNs can be applied to. The recommended network can adapt to changes in the number of input points which is the limit of current methods. Experiments show that in the tasks of classification and part segmentation, PE-Net achieves the state-of-the-art performance in multiple challenging datasets, such as ModelNet and ShapeNetPart.

*Index Terms*— point-based, representation of point clouds, encode

## 1. INTRODUCTION

High precision point clouds can restore the real world and are widely used in the games, construction and self-driving. Due to the limitation of hardware technology, the 3D point clouds acquired are sparse and unordered. [1, 2] apply 3D convolutional neural network to 3D data that is represented by fixed-size voxel grids, which most computations are redundant because of the sparsity of point clouds, and the performance of networks is largely affected by the resolution loss and computational cost. Although 2D convolutional neural networks have been shown to be effective and efficient in dealing with regular data, such as image segmentation, speech recognition and so on, its application to 3D data remains a challenge. A few recent works have studied how to apply deep learning to unordered sets with 2D convolutional neural network, which can be roughly categorized into three main types of pipelines.

**Regular Processing.** [3, 4, 5] obtain training data by projecting point clouds into regular spaces, such as flat or spherical and extract feature through typical 2D convolution. Although some surface information is lost during projection caused by occlusion, and the selection of projection angle is usually artificial, the method still retains high performance.

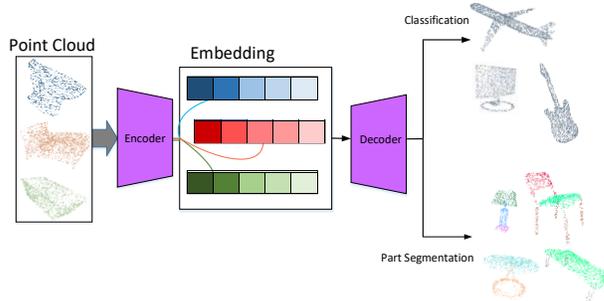

**Fig. 1.** A brief framework of PE-Net and its application. Our method encodes unordered point clouds to high-dimensional regular space, which can be applied to classification and part segmentation with typical 2D CNNs as decoder.

[6, 7, 8] process the input point cloud by using the specific data structures of Octree and KD-Tree, respectively. Li et al. [9] simulate the spatial distribution of point clouds by constructing self-organizing maps (SOM). These methods need a certain preprocessing time for the forward transformation of the network, and the rotation of the point cloud would lead to change in data structure, affecting the final performance.

**Point-based.** PointNet [10] is the first deep neural architecture that directly takes the unordered point cloud as input. [11, 12, 13] etc. use PointNet as a base classification component in 3D object detection. However, PointNet extracts global feature from the point cloud and lacks the ability to capture local feature.

Many works have been done to improve the network in this field. PointNet++ [14] adopts the method of stratified sampling to extract local features at different scales and obtain deep features through the multi-layer network structure. PointCNN [15] proposes *X*-transform, which learns the association features of local points and rearranges them into potential normative order. PointConv [16] proposes a density-weighted convolution operation. Wang et al. [17] introduces a differentiable neural network module EdgeConv, which contains local neighborhood information.

The problem of PointNet series methods is that the surface information extracted from point clouds is incomplete. The collection and sampling methods of 3D data would affect the performance of the models.

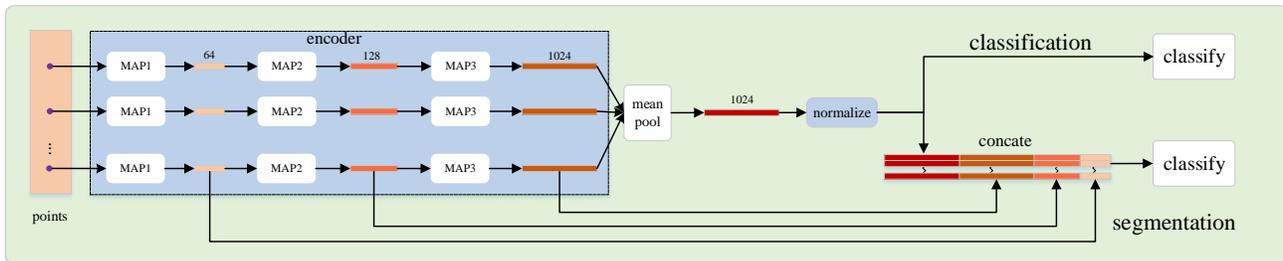

**Fig 2.** The **PE-Net framework** for classification and segmentation. The encoder network has 3 convolutional layers as the mapping network, followed by a mean-pool layer. A normalization layer is utilized to eliminate the effect of the number of points. Typical 2D classification networks are introduced as the head of the architecture.

**Graph Based.** RGCNN [18] treats the characteristics of points in point cloud as signals on a graph using spectral graph theory. They approximate the convolution of the defined graph by Chebyshev polynomials. Wang et al. [19] consider the relative distribution and characteristics of neighborhood points by using spectral convolution and new graph pooling strategy on local graphs. Feng et al. [20] propose HGNN, a hypergraph neural network which can learn the complex spatial representation of point cloud by using the features extracted by [3, 5] as input. HGNN has excellent performance, but it needs great feature extracted by other network as its input, and can't deal with point cloud directly.

In this paper, we propose a network, which can encode points of the input point cloud to a high-dimensional space, and solve the problem caused by the sparsity and disorder of point cloud.

The key contributions of this paper are as follows:
- We design a general deep learning network architecture that directly takes point clouds as input.
- The idea of network is very simple and extensible.
- The unified network architecture can receive different amounts of point clouds as input in inference phase.
- The method performs SOTA on standard point cloud benchmark datasets.

## 2. OUR METHOD

In this section, we analyze the principles of PointNet and propose a novel point-based method to process point clouds.

### 2.1. Rethink PointNet

The attributes of a point cloud are determined by the spatial relationship between points within the point cloud. PointNet extracts the spatial relationship between points through MLP. The input of the network is the whole point cloud. The nodes in each layer of the MLP receive the output from all nodes in the previous layer and process the messages. Therefore, the output information of each node is essentially an abstract description of the spatial relationship between points in the point cloud. The deeper the neural network layer, the more abstract the spatial information output by its nodes. In order to keep the spatial relationship of the output of the middle layer consistent with the order of the input of the previous layer, T-net is used to enforce alignment. At the same time, the spatial relationship that is most suitable for solving task is selected as the feature of the point cloud through the max pooling operation. Therefore, PointNet is essentially a spatial relationship processing network architecture.

Input data with $N$ points has $N!$ kinds of combination. The input order of point cloud has an impact on the result of the network. The number of the input point cloud determines the width of the input layer. If the number of points in point cloud is not constant for each task, it is necessary to sample the points from point clouds, resulting in the information loss of point clouds.

### 2.2. Properties of Point Cloud

In addition to the disorder, sparseness and invariance mentioned in PointNet, point clouds also have the following properties:
- A single point can be regarded as a point cloud with the number of points equal to one.
- Additivity. A point cloud can be regarded as the sum of the point clouds represented by a single point.
- Location uniqueness. Each point in the point cloud occupies a specific position in space, and different points do not overlap with each other.
- Data range consistency. The point clouds obtained by the same sensor share the same data range.

According to the characters of point clouds mentioned above, a point cloud can be represented as a set and can be also represented as the union of subsets which contains all individual points of point clouds. Point cloud $A$ in low-dimensional space $L$ can be represented as follows:

$$A^L = \{p_1, p_3, \ldots, p_n\} = \{p_1\} \cup \{p_2\} \cup \ldots \cup \{p_n\} \quad (1)$$

### 2.3. Point Embedding

The PE-Net structure is built of encoder and decoder network. The encoder is essentially a convolutional layer that trained to map points in point cloud from low-dimensional space $L$ to high-dimensional space $H$. As shown in Fig. 2, the encoder first takes a point as input and maps it from a $1 \times 3$ dimensional vector $p$ into a $1 \times k$ vector $P$. Here, $k$ is a large number. Each bit on the vector $P$ corresponds to a specific position in high-dimensional space $H$, and the value of vector $P$ represents the probability distribution of the input point in high-dimensional space $H$. Then, all high-dimensional vectors are added, and the sum is the distribution of the point cloud in the high-dimensional space $H$. The mapping transformation can be described by

$$f: L \to H \quad (2)$$

where $f$ is the mapping function represented by the encoder network.

| | Representation | Input | ModelNet10 | | ModelNet40 | |
|---|---|---|---|---|---|---|
| | | | Class | Instance | Class | Instance |
| PointNet [2] | points | 1024x3 | - | - | 86.2 | 89.2 |
| PointNet++ [14] | points+normal | 5000x6 | - | - | - | 91.9 |
| Kd-Net [8] | points | $2^{15}$x3 | 93.5 | 94.0 | 88.5 | 91.8 |
| OctNet [7] | octree | $128^3$ | 90.1 | 90.9 | 83.8 | 86.5 |
| O-CNN [6] | octree | $64^3$ | - | - | - | 90.6 |
| SO-Net [26] | points+normal | 5000x6 | 95.5 | **95.7** | 90.8 | **93.4** |
| PE-Net | points+normal | 5000x6 | - | 93.2 | - | 90.8 |

**Table 1.** Object Classification results on ModelNet.

And in high-dimensional space $H$, Point Cloud $A$ can be represented as follows:

$$A^H = P_1 + P_2 + \cdots + P_n \quad (3)$$

The output of the trained encoder network will be in fixed range. This ensures that mapped points are also in the same data space, which satisfies the characteristic that the point cloud collected by the same acquisition device will have the same data range.

We have tested various values of $k$, including 256, 512, 1024 and 2048. Among them, $k = 1024$ achieved best result in this paper. Assuming that each position has only two states of "0" and "1", the $1 \times 1024$ dimensional vector can represent $2^{1024}$ points. In autonomous driving [21], a laser radar with an accuracy of $2cm$ has a scanning space of $80 \times 80 \times 4m$ and can collect up to 1280,000 points ($\approx 2^{20}$). In fact, generally only 100,000 points can be collected. The $1 \times 1024$ dimensional space can fully represent the point cloud space.

Suppose there are $N$ points in the point cloud, and the input batch size is $bs$. For a batch of point clouds, the mapping network needs to undergo $N \times bs$ forward calculations. In order to speed up the calculation, we integrated the inputs as a whole network input. After integration, the input tensor is $m \times 1 \times 1 \times 6$, which only needs one forward calculation, here $m = bs \times N$. The encoder network will output a $m \times 1 \times 1 \times k$ vector. We split the output to a vector of $bs \times N \times 1 \times k$, and sum it in the first dimension to get the vector of $bs \times 1 \times 1 \times k$. This will be the input to the decoder.

Experimental results show that a depth of three is the best choice for the encoder. We will show the results of our tests on different depths of encoder network in Sec 5. The 3-layer neural network maps the point cloud to 1x64, 1x128, and 1x1024 high-dimensional spaces in turn. We can also reshape the point cloud from a 1x1024 vector into a 32x32 two-dimensional vector and use 2D convolution for feature extraction.

For points that belong to the same point cloud, we use the averaging operation to find the distribution vector of the point cloud in the space where it is located. At the same time, in order to maintain the consistency of the feature dimensions of different point clouds, we use max-min normalization for the distribution vector. Experiments show that the normalization operation is conducive to the application of different point cloud distributions on the network.

At the back end of the network, we use a typical neural network structure for feature extraction and task processing. For example, ResNet structure.

## 3. EXPERIMENTS

In this section, the performance of our PE-Net is evaluated in two different applications, namely classification and part segmentation.

### 3.1. Datasets

We evaluated our model on ModelNet dataset, which is a typical benchmark for point cloud classification task. There are two main datasets in ModelNet. ModelNet40 consists of 9,843 training and 2,468 testing annotated objects. 40 categories are defined. ModelNet10 is divided into 3,991 for training set, and 908 for testing set. 10 categories are defined. In this paper, we use the prepared ModelNet10/40 dataset from [1], where each model is sampled by farthest point sampling.

MNIST dataset [22] is a dataset of handwritten digits, which we tested on, to evaluate whether our method can behave like typical 2D classification CNNs. For each digit, 5000 points are randomly sampled from the non-zero pixels as input of the model. A dimension of a little random number is added to each 2D point to from 3D data.

To demonstrate the generality of our method, we also evaluate PE-Net on ShapeNetPart [23], a standard dataset for part segmentation. It contains 16,881 annotated objects, including 12,137 for training, 1,870 for validation and 2,874 for testing. There are 16 categories and total 50 parts in the dataset. In our experiments, we sampled a fixed 1024 points for each point cloud.

To make the model more robust to various input object sizes and shapes, each training object is pre-processed before feeding into the network. The operations that are covered are as follows: zero-mean normalization, randomly jittering, randomly shift and randomly scale. The operation of randomly rotation is not considered because both the training and test sets are aligned.

|  | mean | aero | bag | cap | car | chair | ear. | gui. | knife | lamp | lap. | motor | mug | pistol | rocket | skate | table |
|---|---|---|---|---|---|---|---|---|---|---|---|---|---|---|---|---|---|
| SHAPES(Test) | - | 341 | 14 | 11 | 158 | 704 | 14 | 159 | 80 | 286 | 83 | 51 | 38 | 44 | 12 | 31 | 848 |
| PointNet [10] | 83.7 | 83.4 | 78.7 | 82.5 | 74.9 | 89.6 | 73.0 | 91.5 | 85.9 | 80.8 | 95.3 | 65.2 | 93.0 | 81.2 | 57.9 | 72.8 | 80.6 |
| PointNet++ [19] | **85.1** | 82.4 | 79.0 | 87.7 | 77.3 | 90.8 | 71.8 | 91.0 | 85.9 | 83.7 | 95.3 | 71.6 | 94.1 | 81.3 | 58.7 | 76.4 | 82.6 |
| Kd-Net [14] | 82.3 | 80.1 | 74.6 | 74.3 | 70.3 | 88.6 | 73.5 | 90.2 | 87.2 | 81.0 | 94.9 | 57.4 | 86.7 | 78.1 | 51.8 | 69.9 | 80.3 |
| SONet [9] | 84.6 | 81.9 | 83.5 | 84.8 | 78.1 | 90.8 | 72.2 | 90.1 | 83.6 | 82.3 | 95.2 | 69.3 | 94.2 | 80.0 | 51.6 | 72.1 | 82.6 |
| PE-Net | 83.9 | 82.6 | 78.1 | 83.8 | 73.4 | 89.7 | 71.3 | 90.2 | 85.7 | 79.4 | 94.8 | 64.5 | 87.7 | 77.6 | 58.1 | 70.3 | 83.2 |

**Table 2.** Object part segmentation results on ShapeNetPart dataset.

### 3.2. Classification

The classification accuracy on ModelNet is shown in Table 1. Our proposed PE-Net model achieves a competitive performance. The PE-Net reports a testing instance accuracy of 90.8% on ModelNet40, which is on-par with the O-CNN model, an improvement of 1.6% over 89.2%, achieved by the baseline PointNet. Table 3. illustrates the comparison with the state-of-the-art methods on the MNIST test set. The PE-Net reports a testing accuracy of 99.37%, an improvement of 0.17% over 99.20%, achieved by LeNet. These results firmly demonstrate the effectiveness of PE-Net.

| Method | MNIST |
|---|---|
| LeNet [24] | 99.20 |
| Network in Network [25] | 99.53 |
| PointNet++ [14] | 99.49 |
| PointCNN [15] | 99.54 |
| SO-Net [9] | **99.56** |
| **PE-Net** | 99.37 |

**Table 3.** Image classification results.

**Adapt to change of point number.** We train our network with point clouds of fixed amount, but test it with number change. As shown in Fig. 3, when the number of test points is greater than 256 points, the test performance of each network tends to be stable. Unexpectedly, the best result is not the number of test points equals the number of training points. We found that the overall classification performance generally increased with the number of test points, which is consistent with the intuition that the more the number of points, the more information.

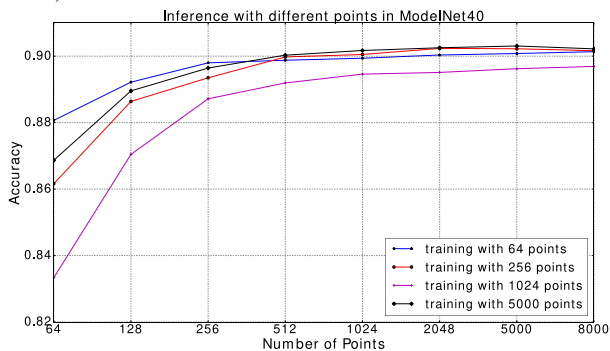

**Fig 3.** Adapt to different number of points

**Ablation Study.** We investigated the effects of the depth of encoder network and the number of training points on the ModelNet40 dataset. As can be seen from Fig 4., the best choice of the number of layers of the encoder network is three, which have achieved good performance for different numbers of input points.

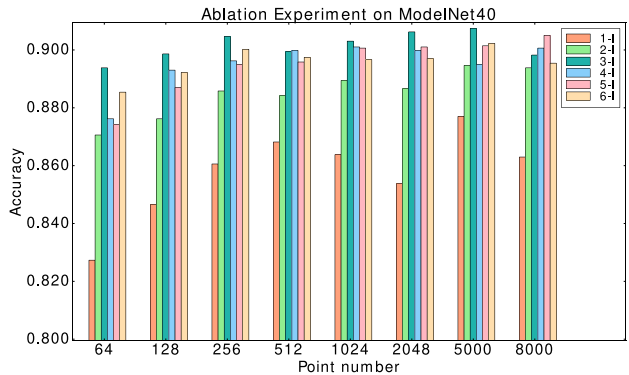

**Fig 4.** Effect of the depth of encoder and the number of input points on classification accuracy with ModelNet40

### 3.3. Part Segmentation

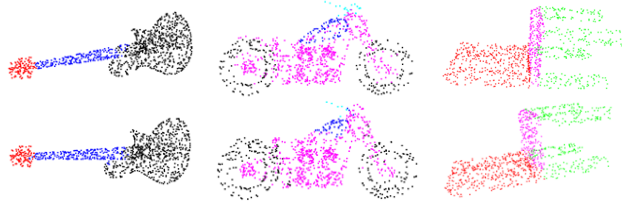

**Fig 5.** Outputs of our network for object part seg: *top* is ground truth; *bottom* is predicted result. From *left* to *right*: guitar, motor, chair.

Our approach addresses the object part segmentation problem by linking each point in an object to a part class label. The results on ShapeNetPart test set are shown in Table 2. The PE-Net model performances competitively with top-tier methods, and out-performs the baseline PointNet by 0.2%. Fig 5. shows the visualization of some segmentation results on ShapeNetPart, which are generally visually satisfying. No pre-processing required and the low-computation cost remains as our advantages.

### 4. CONCLUSION

In this paper, we propose PE-Net, which applies 2D CNN directly to point clouds. Our approach addresses the problem that PointNet family methods require a fixed number of input points. An intuitive explanation of our contribution lies in that we take advantage of the additivity of the probability to get a high-dimensional representation of the point cloud, and use normalization to eliminate the effect of quantity. We believe that our idea of learning point embedding can be potentially generalized to 3D related tasks. We leave as our future work.